%% file: iclr2027_conference.tex
\documentclass{article} 
\PassOptionsToPackage{nonamebreak}{natbib}
\usepackage{iclr2027_conference,times}

\input{math_commands.tex}

\usepackage{hyperref}
\hypersetup{
    colorlinks=true,
    allcolors=black
  }
\usepackage{url}
\usepackage{amsmath,amssymb,amsthm}
\usepackage{booktabs}
\usepackage{array}
\usepackage{siunitx}
\usepackage{graphicx}
\usepackage{placeins}
\usepackage{tikz}
\usetikzlibrary{positioning,fit,backgrounds,arrows.meta}
\usepackage{natbib}
\usepackage{makecell}
\usepackage{tcolorbox}
\usepackage{float}
\usepackage{fontawesome5}
\usepackage{latexml}

\title{FinancialAuditBench: Benchmark Construction under Differential Privacy Using Real-World Priors}

\iflatexml
    \author{%
      Jerry Huang\thanks{\raggedright
      Equal contribution.
      Work completed while working at Modus.
      Correspondence to:
      \href{mailto:jerry8@illinois.edu}
        {\textcolor{blue}{\texttt{jerry8@illinois.edu}}},
      \href{mailto:sarveshb@stanford.edu}
        {\textcolor{blue}{\texttt{sarveshb@stanford.edu}}},
      and \href{mailto:matt.vanburen@modusaudit.com}
        {\textcolor{blue}{\texttt{matt.vanburen@modusaudit.com}}}.}\\
      University of Illinois Urbana-Champaign
      \And
      Sarvesh Babu\textsuperscript{*}\\
      Stanford University
      \And
      Matt Van Buren\\
      Modus Audit Inc.
      \And
      Alexander Wang\\
      Modus Audit Inc.
      \And
      Pranav Pillai\\
      Modus Audit Inc.
      \And
      Arush Jain\\
      Modus Audit Inc.
      \And
      James P. Burton\\
      Modus Audit Inc.
      \And
      Julia Hockenmaier\\
      University of Illinois Urbana-Champaign
    }
  \else
    \author{%
      \expandafter\def\csname @makefnmark\endcsname{}%
      Jerry Huang\textsuperscript{*\,1}%
      \thanks{\raggedright
      Equal contribution.
      Work completed while working at Modus.
      Correspondence to:
      \href{mailto:jerry8@illinois.edu}
        {\textcolor{blue}{\texttt{jerry8@illinois.edu}}},
      \href{mailto:sarveshb@stanford.edu}
        {\textcolor{blue}{\texttt{sarveshb@stanford.edu}}},
      and \href{mailto:matt.vanburen@modusaudit.com}
        {\textcolor{blue}{\texttt{matt.vanburen@modusaudit.com}}}.} \quad
      Sarvesh Babu\textsuperscript{*\,2} \quad
      Matt Van Buren\textsuperscript{3} \quad
      Alexander Wang\textsuperscript{3}
      \\[0.1cm]
      \bfseries Pranav Pillai\textsuperscript{3} \quad
      Arush Jain\textsuperscript{3} \quad
      James P. Burton\textsuperscript{3} \quad
      Julia Hockenmaier\textsuperscript{1}
      \\[0.2cm]
      {\normalfont
        \textsuperscript{1}University of Illinois Urbana-Champaign
        \quad
        \textsuperscript{2}Stanford University
        \quad
        \textsuperscript{3}Modus Audit Inc.}
    }
  \fi

\iclrfinalcopy 
\begin{document}

\maketitle
\lhead{Preprint}

\begin{abstract}
As AI agents are becoming widely adopted in the financial services industry, careful measurement is essential to understand where they can be reliably deployed and where oversight and professional review remain necessary. Such measurement, however, is constrained by limited access to proprietary or privacy-sensitive data. Existing benchmarks therefore often rely on publicly available data, human- and/or LLM-authored tasks, or simplified settings. We introduce FinancialAuditBench, a benchmark for evaluating agents on financial statement audit tasks, along with a framework for systematically generating synthetic engagements. Our task generation framework leverages differentially private aggregate statistics from historical audits along with audit expertise contributed through over 1,100 hours of benchmark development and review. FinancialAuditBench consists of 90 tasks spanning workpaper completion and review across six synthetic audit engagements, each containing an average of 179 files. Evaluation on eleven frontier models shows that while agents complete substantial portions of staff-level audit tasks well, they sometimes perform inappropriate procedures or produce incorrect documentation. Beyond financial auditing, our framework offers an approach for systematically generating synthetic tasks for model evaluation and training in privacy-sensitive domains.
\end{abstract}

\begin{center}
  \begin{tabular}{@{}cll@{}}
      \faIcon[regular]{window-maximize}
  & \textbf{\small Website}
  & \href{https://financialauditbench.com/}
         {financialauditbench.com} \\

      \faDatabase
      & \textbf{\small Dataset}
      & \href{https://huggingface.co/datasets/modusaudit/FinancialAuditBench}
             {huggingface.co/datasets/modusaudit/FinancialAuditBench} \\

      \faGithub
      & \textbf{\small Code}
      & \href{https://github.com/modus-audit/financial-audit-bench}
             {github.com/modus-audit/financial-audit-bench}
  \end{tabular}
  \end{center}

\section{Introduction}
As artificial intelligence capabilities have continued to improve, financial institutions are increasingly adopting generative AI in their workflows, with 71\% of financial services firms having reported adoption in a recent survey~\citep{ccaf2026globalai}. Yet, while this domain is highly suitable for the strong information retrieval and reasoning capabilities that agents have shown to possess, publicly available benchmarks remain limited in number and complexity. As a consequence, researchers and financial experts lack the datasets required to accurately measure where autonomous agents can be deployed reliably and where professional oversight remains necessary.

Due to the privacy-sensitive and confidential nature of financial services, further progress in the evaluation and training of frontier models is increasingly constrained by the amount of accessible data~\citep{pmlr-v235-villalobos24a, muennighoff2023scaling}. Existing approaches to benchmark construction rely heavily on publicly available data~\citep{jimenez2024swebench, islam2023financebenchnewbenchmarkfinancial}, use large language models (LLMs) to propose and validate tasks~\citep{gandhi2026gobrowse, xie2026agentsynth}, and/or require many hours of work by domain experts~\citep{wang2026bigfinancebenchworkflowgroundedbenchmarkfinancialresearch, harvey-legal-bench, xu2026theagentcompany}. To construct \textit{realistic} financial tasks and environments, key challenges remain in leveraging privacy-sensitive data with appropriate privacy guarantees and developing scalable methods for task construction.

\begin{figure}[t]
\centering
\includegraphics[width=\textwidth]{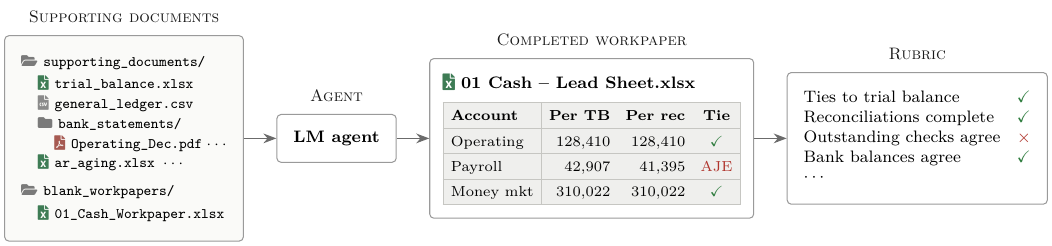}
\caption{Task structure in FinancialAuditBench. The agent uses the provided supporting documents to perform the appropriate audit procedures in a workpaper template. The completed workpaper is then evaluated against a rubric.}
\label{fig:benchmark-overview}
\end{figure}

In this work, we focus on financial statement auditing\footnote{Key audit terminology is defined at the start of Section~\ref{sec:benchmark}.}, a domain that examines whether a company's reported financials are supported by sufficient evidence and are accurately reported within a set of tolerance bounds. Auditors identify areas where significant reporting errors could occur, obtain supporting evidence, and test the company's reported amounts against that evidence. For example, checking reported revenue may require tracing a recorded sale to a customer order, an invoice, and shipping confirmations. Thus, a data room used for evaluation or training requires a set of financial records whose amounts, dates, and relationships are internally consistent across documents.

To solve these aforementioned challenges, we develop a computational graph that generates synthetic companies and their corresponding financial documents. Auditors help define this graph's variables, dependencies, and accounting rules, while differentially private aggregate statistics from historical audits provide priors for a subset of selected variables. These aggregate statistics ground task construction in observed historical patterns with the goal of reducing the amount of domain expert feedback necessary to correct implausible initial assumptions otherwise generated from public sources or from LLMs. Differential privacy furthermore provides formal guarantees for releasing aggregate statistics from confidential data~\citep{roth-dp}, while simpler anonymization methods such as removing personally identifiable information (PII) have been shown to not necessarily prevent sensitive information from being inferred from the remaining context~\citep{oh-etal-2026-subject, 4531148}. Section~\ref{sec:construction} describes the task generation pipeline, its construction process, and the privacy boundaries established during construction, while Section~\ref{sec:dp} presents the formal differential privacy guarantee.

Using this framework, we construct FinancialAuditBench~(Figure~\ref{fig:benchmark-overview}), a benchmark of 90 workpaper-completion and review tasks that measures agents' accuracy and reliability on a set of commonly performed real-world audit tasks, required in a financial statement audit as performed by licensed certified public accountants (CPAs). Each task provides an agent with supporting documents from a synthetic engagement binder and a workpaper to complete or review. The submitted workpapers are then evaluated against rubrics designed by licensed auditors.

We evaluate eleven open- and closed-weight frontier models in a range of realistic settings commonly encountered in practice. While prior work has shown that LLMs perform well on CPA exam questions~\citep{eulerich2024hype, zacher2024cpaexam}, we find that agents still struggle to complete real-world audit tasks reliably without human supervision or assistance. More specifically, the best-performing frontier model achieves a single-attempt task pass rate of 69.17\% (Pass@1), falling to 44.44\% as measured by pass\textasciicircum{}8~\citep{yao2024taubenchbenchmarktoolagentuserinteraction}. Thus, audit work completed by frontier models still requires review by domain experts and, where necessary, correction before it can be used in practice. Overall, we believe the quality, scale, and complexity of FinancialAuditBench make it a realistic and representative benchmark for evaluating agents' abilities to complete common financial auditing procedures.

Our work makes three primary contributions: (1) FinancialAuditBench, a realistic benchmark containing 90 tasks across six synthetic audit engagements, (2) a systematic framework for generating internally consistent synthetic audit engagements, and (3) an empirical evaluation of eleven frontier models across several experimental settings.

We release a development version of the task generation framework under the MIT license for use by the broader research community. 

\section{Related Work}
\paragraph{Benchmarks for AI systems.} Benchmarks for agent evaluation have progressed from information retrieval tasks to longer-horizon, real-world tasks. While initial benchmarks focused on simpler question-answering and general tool-use settings~\citep{islam2023financebenchnewbenchmarkfinancial, patil2025the}, more recent benchmarks have focused on knowledge work settings that more closely mirror work that professionals do~\citep{lee-etal-2026-evaluating}. Within the domain of finance, APEX–Accounting~\citep{benchek2026apexaccounting} evaluates accounting workflows such as recording transactions, reconciling accounts, and preparing reports. AuditFraudBench~\citep{liu2026auditfraudbenchbenchmarkingauditjudgment}, on the other hand, evaluates audit judgment using company filings and regulatory evidence. FinancialAuditBench focuses on financial statement audit fieldwork, requiring agents to examine engagement evidence, perform specified audit procedures, and document their findings in workpapers.

\paragraph{Scalable task construction.} Several prior works have proposed methods for systematic task generation rather than relying solely on manual task creation. In software engineering, SWE-smith~\citep{yang2026swesmith} uses an automated pipeline to construct tasks and write tests to validate candidate tasks. In the domain of computer-use, AgentSynth~\citep{xie2026agentsynth} uses LLM-based task proposal, execution, and verification to construct long-horizon tasks, and Go-Browse~\citep{gandhi2026gobrowse} uses LLMs to propose webpage-grounded tasks through structured exploration. In finance, FinMaster~\citep{jiang2025finmasterholisticbenchmarkmastering} uses a financial simulator to generate transactions and financial statements, with auditing tasks focusing on identifying injected errors in transaction records.

\paragraph{Differential privacy.} Prior work has developed differentially private generative models, including GANs~\citep{yoon2018pategan} and diffusion models~\citep{dockhorn2023differentially}. For generating data, AIM~\citep{mckenna2024aimadaptiveiterativemechanism} generates synthetic tabular data from privately measured marginals, and Private Evolution~\citep{lin2025dpsdaimages} and Sim-PE~\citep{lin2025dpsdasimulators} use differentially private nearest-neighbor counts to guide iterative generation with foundation models and simulators, respectively. Other approaches use LLMs to generate differentially private synthetic text~\citep{amin-etal-2024-private} or generate differentially private examples for in-context learning~\citep{gao2025dataadaptive}.

\section{FinancialAuditBench}
\label{sec:benchmark}
\paragraph{Terminology.} An \textit{engagement} is the audit of a company for a specific reporting period, usually a fiscal year. \textit{Prepared-by-client} (PBC) files or PBCs are records, financial documents, and ledgers supplied by the audited company or client. In FinancialAuditBench, a \emph{workpaper} is a spreadsheet organized using a predefined template in which auditors document procedures performed, evidence examined, and conclusions reached. An engagement \textit{binder} contains the materials for one synthetic audit engagement, including planning documents, client-provided records, third-party evidence, and blank workpaper templates. Agents use these materials to produce completed workpapers.

\paragraph{Overview.} FinancialAuditBench is a benchmark dataset that evaluates how well modern AI systems complete and review real-world fieldwork tasks typically performed by CPAs during a financial statement audit. The benchmark includes financial documents for six synthetic companies spanning two industries, manufacturing and staffing services. Appendix~\ref{app:binder-structure} describes the types of files in an example engagement binder. Each synthetic company's binder includes either seven or eight workpapers, depending on industry, that can be independently completed, where a single task corresponds to completing or reviewing one workpaper. In aggregate, FinancialAuditBench comprises 45 main tasks that measure workpaper completion performance and 45 corresponding review tasks.

The tasks in FinancialAuditBench are designed to model work typically assigned to more junior (staff-level) auditors during the substantive fieldwork phase of a financial statement audit. Agents receive an audit plan and assessed risks as inputs, and the workpaper templates specify the procedures to perform. Completing these tasks requires identifying and evaluating relevant evidence, checking recorded amounts against supporting documents, performing calculations and reconciliations, and documenting supported conclusions. The synthetic client binders, moreover, are designed to provide much of the supporting documentation required to complete these procedures, while limiting common errors and ambiguities in the supporting information (although some remain as discussed in Appendix~\ref{app:audit-methodology}). Additional information on how these tasks relate to financial audits in practice can be found in the appendix as well.

\paragraph{Task format.} As illustrated in Figure~\ref{fig:benchmark-overview}, each task provides an agent with a synthetic client's set of supporting documents and one workpaper template specifying the audit procedures to perform. Similar to how auditors complete these workpapers, the agent must examine the provided files and perform the specified audit procedures. We release all the tasks in the Harbor format~\citep{Harbor_Framework}, which packages task instructions, execution environments, and grading scripts.

\paragraph{Evaluation.} Completed workpapers are graded against rubrics covering an auditor-guided sample of data points, calculations, and conclusions, using reference answers prepared and/or reviewed by CPA-licensed auditors. This grading approach is motivated by traditional workpaper review in audit firms, where reviewers exercise judgment about what to examine rather than systematically reperforming every procedure or checking every cell. The rubrics contain an average of ten scored checks per workpaper, and a workpaper receives a passing score when all applicable criteria are met; otherwise, it fails under a strict all-or-nothing grading metric. Importantly, a passing score under our proposed evaluation methodology \textbf{does not imply that a given workpaper is free of errors or omissions outside of the rubric's scope}. The percentage of criteria satisfied as a partial-credit measure of performance on the selected checks is also provided. Additional details on how evaluation and grading criteria are implemented can be found in Appendix~\ref{app:experimental-details}.

\section{Benchmark Construction}
\label{sec:construction}

\subsection{Preliminaries}
\paragraph{Privacy boundary.} Before constructing the benchmark, we establish explicit rules governing access to privacy-sensitive historical audit data. Only Phase II, as outlined below, accesses confidential source binders. Later stages use the resulting differentially private statistics, public information, and general audit expertise. Auditors provide feedback on synthetic materials in environments that prohibit access to confidential source documents and are instructed to not reference any specific engagements they have worked on in the past.

\paragraph{Framework structure.} An engagement binder contains over a hundred financial documents that describe snapshots of the same shared underlying business activities. Their amounts, dates, and relationships must therefore remain consistent across documents. To enforce these constraints, the framework represents each generated company as a computational graph whose nodes encode company characteristics and business events, and whose edges specify the relationships among them.

At a high level, the graph contains three types of nodes. Source nodes encode user-defined inputs or values sampled from predefined distributions, conditionally sampled nodes draw values from distributions determined by the values of their parent node(s), and deterministic nodes apply accounting rules or calculations. Figure~\ref{fig:computational-graph} below illustrates these dependencies with a simple payroll example. Here, the chosen industry determines the distributions from which annual revenue and the payroll-to-revenue ratio are sampled. Multiplying the sampled revenue by the sampled ratio then determines annual payroll. This design represents dependencies across financial records and alleviates the need for estimating high-dimensional joint distributions.

\subsection{Benchmark Scope}
We first defined the benchmark's scope as engagement binders spanning the industries of \textit{manufacturing} and \textit{staffing services}. Following these decisions, we consulted fifteen auditors to identify the information used when auditing these industries including client-provided records, third-party evidence, auditor-developed materials, and public information. The final benchmark includes eight workpaper types, with supporting information identified through auditor feedback.

\subsection{Framework Construction}
We construct the task generation framework in four phases. For each phase, we clearly outline where and how LLMs were used. Additional implementation details can be found in our released code.

\begin{figure}[t]
\centering
\includegraphics[width=0.85\textwidth]{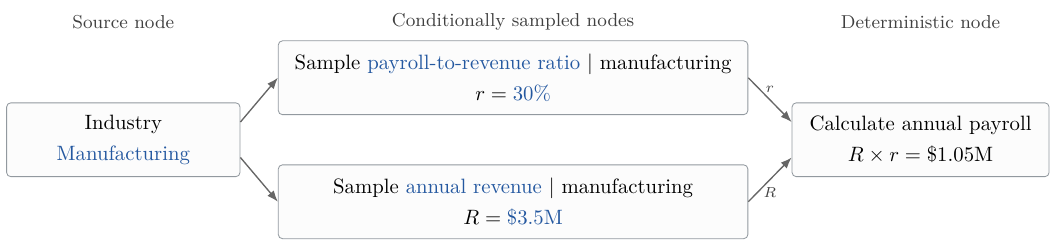}
\caption{A toy example illustrating the three node types used in the task generation graph. The source node specifies the industry, which determines the distributions from which annual revenue and the payroll-to-revenue ratio are sampled. The deterministic node calculates annual payroll by multiplying the two sampled values.}
\label{fig:computational-graph}
\end{figure}

\paragraph{Phase I: Variable selection.} Using the workpaper types and supporting document categories identified during scoping, we create document drafts with input from domain experts and AI tools, refining their realism, accuracy, and presentation based on auditor feedback. At this stage, these drafts specify the content each document should contain but do not yet capture the relationships across documents needed to form a coherent engagement binder. An LLM then proposes candidate variables from these drafts, and three auditors independently assess each variable from its description and propose additions. A senior auditor, defined in this work as an auditor with over ten years of audit experience, consolidates their feedback, which an LLM then uses to rank the variables for subsequent stages.

\paragraph{Phase II: Private aggregation.} An LLM extracts values for the selected variables from a corpus of real-world historical binders, containing both supporting documents and filled out workpapers. Each binder’s contribution to each statistic is bounded, and the contributions are aggregated across binders using differentially private mechanisms. These statistics provide priors for graph construction and synthetic data generation. Section~\ref{sec:dp} describes the mechanisms and privacy guarantee. In initial construction experiments, we qualitatively found that these empirical priors provide a better basis for constructing the graph compared to those generated by an LLM or derived from publicly available data. In practice, these priors help reduce the corrective feedback required from auditors regarding implausible values and relationships in the generated documents during expert review (Phase IV). Furthermore, because the computational graph models intricate details of each synthetic company (e.g., the proportion of raw material wasted during production at a mid-sized manufacturing company), we found that auditors and financial experts alike could not always confidently assess whether these modeled values were realistic without being able to consult reference documents containing privacy-sensitive data.

\paragraph{Phase III: Graph construction.} To scale the graph, we add edges and derived nodes to the computational graph using public resources on finance and auditing, auditor expertise, and LLMs. The fully constructed computational graph contains around 250 nodes and 1,000 edges. The generator samples values from the differentially private priors and propagates them through the computational graph to derive account counts, balances, financial events, etc. LLM-generated tests check the feasibility of the generated financials and optionally trigger resamples when infeasible company states are reached (e.g., month-end balances that are negative).

\paragraph{Phase IV: Expert review and revisions.} In each round, the generator produces new binders, which auditors review for factual inconsistencies, implausible content, formatting issues, and missing documents. We use this feedback to revise the graph by resolving inconsistencies, modeling additional variables, and fixing presentation issues, then regenerate the affected documents for further review.

\subsection{Human Annotation}
\paragraph{Ground truth and rubrics.}
To construct the ground truth used in FinancialAuditBench, at least one CPA-licensed auditor completed and/or reviewed each workpaper. A senior auditor reviewed the proposed ground truth answers, adjudicated disagreements, and corrected errors. To develop the rubrics, a senior auditor constructed example rubrics for a set of core audit areas and then provided that example rubric structure to an AI model to replicate across the rest of the benchmark tasks. Subsequently, the rubrics were checked and revised by CPAs against multiple model submissions to assess their fairness and robustness.

\paragraph{Human annotation hours.} Across graph construction, task generation, ground-truth preparation, and rubric refinement, auditors contributed a little over 1,100 hours. We do not report any inter-annotator agreement metrics~\citep{cohen1960coefficient} because much of the feedback was developed collaboratively in discussions and collected across multiple disjoint revision cycles. For workpaper template construction, rubric construction and review, and workpaper completion, all final work products were reviewed by a senior auditor.

\section{Differential Privacy}
\label{sec:dp}
\paragraph{Definition and privacy unit.}
The privacy unit is a single historical audit binder. Let \(D\) be a dataset of records from the set of historical audit binders we use in this project. We use binder-level adjacency, in which two datasets \(D\) and \(D'\) are adjacent if one can be obtained from the other by adding or removing all records from a single binder. A randomized mechanism \(M\) satisfies \((\varepsilon,\delta)\)-differential privacy if, for every pair of adjacent datasets \(D,D'\) and every measurable output set \(S\)~\citep{roth-dp},
\[
\Pr[M(D)\in S]
\leq e^\varepsilon \Pr[M(D')\in S] + \delta.
\]

\paragraph{Contribution bounds and mechanisms.}
Our privacy accounting covers six financial ratios, sixteen features describing financial records, and summary statistics for adjusting journal entries and accounts payable activity. Histograms and categorical distributions are estimated by summing bounded contributions from individual binders and applying the discrete Gaussian mechanism~\citep{canonne2020discrete} for ease of accounting under zero-concentrated differential privacy (zCDP). Other mechanisms such as discrete Laplace noise~\citep{ghosh2009universallyutilitymaximizingprivacymechanisms} could also be used for these queries. For each financial ratio, a fixed-grid variant of JointExp~\citep{gillenwater2021quantiles} jointly estimates seven quantiles. Appendix~\ref{app:privacy} details the contribution bounds and privacy accounting, and our released code provides additional implementation detail.

\paragraph{Privacy guarantee.}
Before accessing any historical binders, we set a binder-level privacy ceiling after assessing with auditors the disclosure risks associated with the set of selected features as well as the theoretical upper bounds on binder membership inference. We did not empirically optimize this budget as similar data was not available in the public domain to be used for calibration.

Thus, composing the aforementioned mechanisms gives a total zCDP charge of \(\rho\approx0.2505\)~\citep{bun2016zcdp}. Converting this budget at \(\delta=10^{-5}\) gives \(\varepsilon\approx3.647\), yielding a \((3.65,10^{-5})\)-DP guarantee for the addition or removal of one binder, below the predetermined ceiling. Subsequent graph construction and synthetic data generation use only these released statistics, public information, and general audit expertise, and therefore incur no additional privacy loss~\citep{roth-dp}.

\section{Experiments}
\label{sec:experiments}
We evaluate FinancialAuditBench across eleven frontier models on the 45 completion tasks, with eight independent runs per model-task pair. Thus, a total of 3,960 trajectories are generated and released for the main experiments in this work. We additionally evaluate ten models on the 45 review tasks, with one run per model-task pair, yielding an additional 450 review trajectories.

\subsection{Models and Agent Environment}
We evaluate all eleven models in a simple agent harness where the only tool provided to the agent is a \textit{bash} tool, which executes shell commands. Table~\ref{tab:model-configurations} lists the models, API names, reasoning effort, and temperature settings. Each trajectory is limited to 200 model calls. Additional details on the experimental setting are provided in Appendix~\ref{app:experimental-details}.

\subsection{Main Results}
\begin{figure}[!htbp]
\centering
\includegraphics[width=0.8\textwidth]{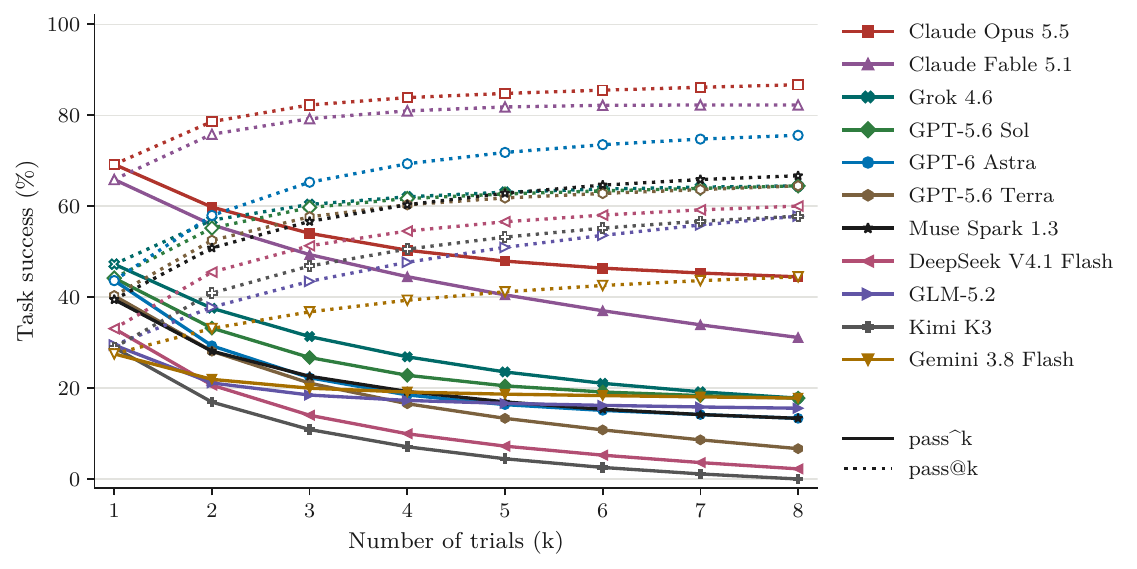}
\caption{Reliability across eight independent runs. Solid lines show pass\textasciicircum{}k and dotted lines show pass@k. Both metrics use strict task success and are averaged across tasks. Exact values are provided in Appendix~\ref{app:reliability-results}.}
\label{fig:reliability}
\end{figure}

Our main experiments reported in Table~\ref{tab:main-results} evaluate \textit{how well AI agents can autonomously complete common audit tasks end-to-end.} We report the task pass rates, partial-credit scores, cost, and runtime across eight independent runs per model-task pair. Strict pass rates measure whether all applicable rubric criteria are satisfied, while partial-credit scores indicate the proportion of criteria satisfied. Additionally, Figure~\ref{fig:reliability} shows pass\textasciicircum{}k~\citep{yao2024taubenchbenchmarktoolagentuserinteraction} and pass@k metrics to measure the reliability of these models across multiple trials. Tables~\ref{tab:pass-k} and~\ref{tab:pass-at-k} in Appendix~\ref{app:reliability-results} report the underlying values.

Across all models, we find that while agents achieve partial-credit scores all greater than 85\%, the best-performing models, Claude Opus 5.5 and Claude Fable 5.1, produce work that satisfies all applicable rubric criteria 69.17\% and 65.83\% of the time, respectively. Furthermore, the pass\textasciicircum{}k results demonstrate limited reliability of these agents across multiple runs with Claude Opus 5.5 and Claude Fable 5.1 passing all eight attempts on only 44.44\% and 31.11\% of tasks, respectively. Thus, these results suggest that deploying agents into financial audit workflows without professional oversight is ill-advised. 

\begin{table}[h!]
\centering
\caption{Workpaper-completion performance across frontier models. Strict pass rates and partial-credit scores are averaged across tasks and runs. Cost and runtime are means per trajectory.}
\label{tab:main-results}
\small
\setlength{\tabcolsep}{5pt}
\renewcommand{\arraystretch}{1.08}
\begin{tabular}{@{}lc*{4}{S[table-format=2.2]}@{}}
\toprule
Model & Weights & \multicolumn{1}{c}{\makecell{Strict pass\\(\%)}} & \multicolumn{1}{c}{\makecell{Partial credit\\(\%)}} & \multicolumn{1}{c}{\makecell{Mean cost\\(USD)}} & \multicolumn{1}{c}{\makecell{Mean time\\(min)}} \\
\midrule
Claude Opus 5.5 & Closed & 69.17 & 95.84 & 3.30 & 11.21 \\
Claude Fable 5.1 & Closed & 65.83 & 94.81 & 6.49 & 13.03 \\
Grok 4.6 & Closed & 47.22 & 91.12 & 1.62 & 11.99 \\
GPT-5.6 Sol & Closed & 44.17 & 89.41 & 0.91 & 3.78 \\
GPT-6 Astra & Closed & 43.61 & 90.95 & 2.96 & 8.00 \\
GPT-5.6 Terra & Closed & 40.28 & 88.66 & 0.40 & 2.06 \\
Muse Spark 1.3 & Closed & 39.44 & 89.62 & 1.87 & 6.64 \\
DeepSeek V4.1 Flash & Open & 33.06 & 88.31 & 0.14 & 16.05 \\
GLM-5.2 & Closed & 29.44 & 87.56 & 0.63 & 12.49 \\
Kimi K3 & Open & 28.89 & 86.01 & 0.63 & 9.35 \\
Gemini 3.8 Flash & Closed & 27.50 & 87.88 & 1.80 & 18.52 \\
\bottomrule
\end{tabular}
\end{table}

\subsection{Failure Analysis}
To understand where and why models struggled and assess the fairness of the rubrics, auditors compared workpapers completed by agents with reference workpapers completed by auditors and reperformed a limited subset of procedures. This qualitative review identified four recurring types of errors: (1) failure to consider the effects of changes, (2) insufficient retrieval, (3) inadequate evaluation of evidence sources, and (4) reluctance to leave items open.

Failure to consider the effects of changes included cases where agents added information to the workpaper without assessing its downstream impacts (e.g., adding numeric details to a table when they should not have been included in the corresponding column total). Insufficient retrieval included cases where agents analyzed only the most obvious source document and concluded, based on that document alone, that the required details were not available (e.g., reviewing a custom shipping cutoff schedule that lacked the necessary detail, while the full shipping log contained it). Inadequate evaluation of evidence sources included cases where agents vouched amounts to client-provided documents rather than third-party evidence (e.g., vouching receivables to a client-prepared cash receipts report instead of a bank statement). Finally, reluctance to leave items open included cases where agents documented work using whatever information was available, even when the information the procedure required was missing, rather than noting that the procedure could not be performed as designed (e.g., documenting work over current-year transactions for a subsequent events procedure when the subsequent-period information had not been provided). These categories were not mutually exclusive.

\subsection{Comparison with Estimated Auditor Completion Times}
Across workpaper areas and all evaluated models, median agent runtimes range from seven to twelve minutes. In comparison, the estimated manual completion times of auditors used in this project ranged between two to twelve hours, when provided the same tasks. These estimates \textit{exclude} senior review and corrective rework and therefore do not establish equivalent time savings in practice. However, they suggest potential time savings when agents are deployed within a well-defined scope and under appropriate professional oversight. Appendix~\ref{app:auditor-time} provides a more in-depth comparison and outline of our time estimation methodology.

\subsection{Review Tasks}
FinancialAuditBench includes 45 review tasks evaluated across ten models, with all models receiving the same starting workpaper and supporting files for each task. In this experimental setting, models are instructed to identify and correct any potential errors that may be present in a filled out workpaper, while preserving correct work. Across the board, all models improve average strict-pass and partial-credit scores, although some individual workpapers received lower scores after review and correction. Claude Opus 5.5 achieves the largest improvement, increasing strict pass from 28.89\% to 62.22\% and partial credit from 87.34\% to 94.57\%. These results suggest that agent review can improve workpaper quality but does not reliably resolve all errors captured by the rubrics. Full results and the review setup are provided in Appendix~\ref{app:review-results}.

\subsection{Computer-Use Setting}
Audit workflows often involve enterprise applications, settings where agents cannot directly interact with applications programmatically and must instead interact through a graphical interface. To explore this setting, we evaluate ChatGPT Computer-Use~\citep{openai2026computeruse} on eight completion tasks from one synthetic binder. The ChatGPT desktop application is granted computer access and instructed to complete all eight workpapers primarily via mouse and keyboard inputs. ChatGPT Computer-Use achieves 92.84\% average partial-credit and a 50.00\% strict pass rate, compared with 93.58\% and 54.69\%, respectively, for a simple agent equipped with a single bash tool on the same tasks. Both experimental settings used GPT-6 Astra. Appendix~\ref{app:experimental-details} provides the computer-use instructions and additional observations.

\section{Discussion}
\paragraph{Implications.} Our results show that while frontier models can complete substantial portions of audit workpapers, they do not yet produce work that can be used without professional review. Given the scope and granularity of our rubrics, the difference between partial-credit and all-or-nothing scoring highlights that models often satisfy many rubric criteria while failing others. Broader and more granular rubric coverage may reveal additional errors that are not reflected in the current scores. In financial auditing, such errors can have substantial real-world consequences if not caught. At the same time, agents help complete these tasks considerably faster than auditors, suggesting a role for AI agents as co-pilots that help auditors perform audit procedures and document their findings under professional oversight.

The value of such a co-pilot role depends not only on the accuracy of agent work, but also on the ease with which auditors can review it. We observed that agents tended to over-document how they arrived at an answer, leaving what the evidence indicated and how it affected the conclusion less clearly stated. Agents also favored writing in sentence fragments and formatted cell content poorly, which increased the effort required for a reviewer to understand what was written. Because the ability to easily understand the procedures performed, the evidence obtained, and the conclusions reached is a core element of quality audit documentation, these tendencies can detract from the overall quality of the work. As a result, such issues may reduce some of the time savings described above. Future evaluations should assess the clarity of agent documentation alongside its correctness.

\paragraph{Limitations and future work.} FinancialAuditBench captures only part of the financial audit process. The current benchmark covers a limited number of industries, companies, workpaper types, and audit methodologies, and its synthetic engagements may not represent the full complexity of actual client work. Future work should expand the benchmark to include additional industries and audit procedures, and seek to model a wider range of realistic conditions, including incomplete, inconsistent, and poorly organized client data. Furthermore, the process of interacting with clients over the course of a financial audit is inherently a multi-turn process with auditors requesting information from multiple counterparties over the course of weeks. Extending this benchmark to support multi-turn conversational task settings would also increase the realism of the benchmark.

Our rubrics evaluate selected aspects of each workpaper and may miss errors or omissions that occur elsewhere. We emphasize that the reported metrics in this work are sufficient to reveal existing limitations in frontier model capabilities, and we \textit{do not} make any claims regarding their readiness for autonomous deployment in real audit engagements beyond the evaluation scope we performed. Systematic and manual construction of more granular rubrics and evaluation methods extends beyond the scope of this paper and remain exciting directions for future work.

Evaluating the realism and fidelity of the generated synthetic binders also remains an open challenge, requiring substantial expert feedback. Despite extensive auditor review, resource constraints and the scale of the generated binders limited our ability to identify and resolve every issue affecting task accuracy and realism. To address these limitations, we restrict evaluation to workpaper sections that have well-defined ground truth, focusing on procedures that are common across financial statement audits in multiple industries.

\paragraph{Conclusion.} We introduce FinancialAuditBench, a benchmark for evaluating AI agents on financial audit tasks, together with a framework for generating synthetic engagements from differentially private aggregate statistics and auditor expertise. Our results show that while agents perform well on many evaluated aspects of audit workpapers, they still do not yet consistently produce workpapers that are sufficient to support an audit conclusion without corrective feedback. These findings suggest that agents can assist auditors with workpaper preparation and review, while professional oversight remains necessary. More broadly, our construction framework shows how realistic tasks in privacy-sensitive domains can be generated for model training and evaluation.

\section*{Acknowledgments}
We thank Jinu Lee and Rishi Bommasani for their helpful feedback at various stages of this work. We also thank Modus Audit Inc. for supporting this research.

\section*{AI Use Statement}
We used generative AI tools spanning AI coding agents and chatbots throughout the development of this project. In early stages of the work, authors of this work used AI tools to learn and better understand the financial audit domain and to more effectively communicate with domain experts. The use of AI tools during the construction of the benchmark itself and its corresponding task generation framework is clearly stated in Section~\ref{sec:construction}. AI tools were also used for creating research figures and tables and for writing parts of the experimental code. The authors inspected and tested AI-assisted code and verified the quality and accuracy of the generated code. We take full responsibility for the final content of this work.

\section*{Ethics Statement}
The benchmark contains synthetic audit records. Historical client data was used with authorization during the construction of this benchmark. The aggregate statistics released are protected by differential privacy bounds as described in Section~\ref{sec:dp}. FinancialAuditBench is intended strictly for research evaluation and should not be used to provide audit assurance or replace professional judgment.

\section*{Reproducibility Statement}
We release the tasks in our proposed benchmark including the data rooms, workpaper templates, rubrics, and relevant code necessary to run the core experiments in this work. For the ChatGPT Computer-Use setting, we provide sufficient detail for other researchers to reproduce our reported results. For the task generation framework, we release a development version of the framework that maintains the final structure of the computational graph used to generate the tasks presented in this paper, but omit substantial domain-specific rules and logic for generating more realistic and complex financial documents. This choice was made so that the code released to the public would be more easily understandable. We publish the query definitions, privacy ledger, and DP-derived priors used by the public generator.

\bibliography{iclr2027_conference}
\bibliographystyle{iclr2027_conference}

\appendix

\section{Audit Task Design, Scope, and Disclosures}
\label{app:audit-methodology}
\subsection{Task Design from an Auditor's Perspective}
The overall process of a financial statement audit spans the general phases of planning, risk assessment, substantive fieldwork, concluding, and reporting, with significant time and effort spent on substantive fieldwork in most audits. Audit evidence gathered during this phase serves as the foundation for the audit opinion and report. The specific class of tasks contained in FinancialAuditBench targets work done during the substantive fieldwork phase, where auditors respond to assessed risks and gather evidence through the performance of audit procedures that produce audit documentation typically in the form of workpapers. Because the planning and risk assessment phases determine the nature, timing, and extent of audit procedures and generally precede substantive fieldwork, the agent is provided with determinations made earlier in the audit, including the assessed risks of material misstatement and the audit plan (i.e., the set of chosen procedures to complete), along with necessary information on the entity and its environment. Given these facts, we assess how accurately an AI system can evaluate the provided evidence, complete the assigned procedures, and reach a well-documented conclusion. 

\subsection{Client-Provided Records and Third-Party Evidence}
In real-world settings, files required for a financial audit are provided either directly by the client or by third-party sources such as banks and the client's customers. These files may include extraneous documents, contain errors, and/or omit essential information. Thus, an auditor must identify the necessary files, request them from the relevant parties, and follow up to resolve errors, omissions, and inconsistencies. Each task in FinancialAuditBench provides the agent with a data room of evidence files and does not permit requests for additional information from clients or third parties. Importantly, as in real-world audits, the provided engagement binders contain imperfections, such as inappropriate account coding and inconsistencies between synthetic documents. Our rubrics do not evaluate parts of each workpaper affected by missing or imperfect information so that agents are not penalized for those limitations.

\subsection{Relationship to Production Audit Practice}
The blank workpaper templates, outlined procedures, documentation conventions, and grading criteria we release in FinancialAuditBench were developed with the intent of evaluating how well AI systems perform on real-world audit tasks. These artifacts are not copies of proprietary production templates and do not use the exact methodology of any accounting firm. Each firm organizes its workpapers differently and uses its own respective testing procedures, review processes, documentation processes, etc. that comply with applicable generally accepted auditing standards. We believe that measured performance on FinancialAuditBench correlates with the performance in production audit settings.

\section{Example Generated Binder Structure}
\label{app:binder-structure}
Figure~\ref{fig:binder-tree} shows a pruned directory tree for an example of a synthetic client binders presented in FinancialAuditBench. At a high level, the planning folder provides assumptions and scope details for the planned audit, the PBC package contains documents prepared by and provided by the client, the direct-to-auditor folder contains third-party evidence such as bank confirmations, and the workpaper templates define the audit procedures that must be performed for a given task.

\begin{figure}[h]
\centering
\begin{minipage}{0.94\textwidth}
\scriptsize
\begin{verbatim}
manufacturing-binder/
|-- planning/
|   |-- Engagement Scope and Case Facts.pdf
|   |-- Materiality Memo.pdf
|   |-- Assertion-Level Risk Assessment.pdf
|   |-- Audit Selection Plan.pdf
|   `-- Planned Procedure Register.pdf
|-- pbc_package/
|   |-- Package Index.xlsx
|   |-- foundational/
|   |   |-- Chart of Accounts.xlsx
|   |   |-- TB 12 31 24.xlsx
|   |   `-- TB 12 31 25.xlsx
|   |-- accounts_receivable/
|   |   |-- Detailed AR aging.xlsx
|   |   `-- Subsequent cash receipts.xlsx
|   |-- ap_and_accruals/
|   |   |-- AP Aging (2025-12-31).xlsx
|   |   `-- Vendor Invoice Support 2025.xlsx
|   |-- inventory/
|   |   |-- Detailed inventory listing or perpetual records.xlsx
|   |   `-- Physical count instructions and results.xlsx
|   |-- revenue/
|   |   |-- Sales invoices and customer support.xlsx
|   |   `-- Shipment delivery or acceptance evidence.xlsx
|   `-- documents/
|       |-- Complete Journal Entry Population.xlsx
|       |-- Payroll Register 2025.pdf
|       `-- Audited Financial Statements 12.31.2024.pdf
|-- direct_to_auditor/
|   |-- Bank Confirmation 12.31.2025.pdf
|   |-- Loan Confirmation 2025.pdf
|   `-- AR Confirmation - [Synthetic Customer].pdf
`-- workpaper_templates/
    |-- cash/cash_workpaper.xlsx
    |-- inventory/inventory_costing_valuation_workpaper.xlsx
    |-- payroll/payroll_workpaper.xlsx
    `-- revenue/revenue_workpaper.xlsx
\end{verbatim}
\end{minipage}
\caption{A snapshot of an example data room provided to agents and auditors for filling out workpaper templates.}
\label{fig:binder-tree}
\end{figure}

\section{Privacy Accounting}
\label{app:privacy}
Each historical binder contributes at most one vote per histogram or categorical query, giving \(\ell_2\)-sensitivity at most one. For each structured summary, the proportions of extracted records in different categories are encoded as a list of weights whose sum is at most 60. For example, if a binder's adjusting journal entries comprise 50\% depreciation adjustments, 25\% amortization adjustments, and 25\% tax adjustments, the corresponding weights would be 30, 15, and 15, with zero weight for all other adjustment types. These weights represent category proportions scaled to 60. Bounding their sum caps the \(\ell_2\)-sensitivity of each structured summary query to 60. 

Using standard mechanism bounds~\citep{canonne2020discrete,cesar2021bounding}, the 36 mechanisms in Table~\ref{tab:privacy-accounting} add up to a total zCDP charge of \(\rho_{\mathrm{total}}\approx0.2505\). Then, standard zCDP-to-DP conversion at \(\delta=10^{-5}\) yields a \((3.65,10^{-5})\)-DP guarantee~\citep{bun2016zcdp}. The public repository includes the query definitions, mechanism parameters, and privacy charges.

\begin{table}[h]
\centering
\caption{Privacy accounting by query family. DG denotes the discrete Gaussian mechanism.}
\label{tab:privacy-accounting}
\small
\begin{tabular}{lc*{2}{S[table-format=1.0]}S[table-format=1.5]}
\toprule
Query family & \multicolumn{1}{c}{\makecell{DG\\\(\Delta_2\leq1\)}} & \multicolumn{1}{c}{\makecell{DG\\\(\Delta_2\leq60\)}} & \multicolumn{1}{c}{JointExp} & \multicolumn{1}{c}{\(\rho\)} \\
\midrule
Financial ratios & 6 & 0 & 6 & 0.04944 \\
Record features & 16 & 3 & 0 & 0.14903 \\
AJE and AP summaries & 2 & 3 & 0 & 0.05203 \\
\midrule
Total & 24 & 6 & 6 & 0.25051 \\
\bottomrule
\end{tabular}
\end{table}

\section{Additional Results and Construction Details}
\label{app:experimental-details}

\subsection{Comparison with Estimated Auditor Completion Times}
\label{app:auditor-time}
Table~\ref{tab:time} compares auditor time estimates with agent performance across eleven models. For each workpaper area, we report the average pass rate and median agent runtime across all models and runs in the main experiments with runtime rounded to the nearest minute. \textit{Auditor completion times are approximate estimates informed by the time auditors spent completing the benchmark workpapers without the use of AI or automation tools.} As mentioned before, based on the definition and structure of each of these tasks, these estimates exclude waiting for client responses and external confirmations as well as senior review and any resulting rework that often arise in actual financial audits. However, auditors' reviews suggested that workpapers with lower partial-credit scores generally would require more corrective feedback and time to fix. Resource constraints limited the number of auditor-completed workpapers, and participating auditors varied in experience. Thus, we do not report an aggregate human accuracy estimate.

\begin{table}[!htbp]
  \centering
  \caption{Estimated manual completion times (without the use of any AI or automation
  tools), measured agent runtimes, and agent pass rates for the benchmark
  workpapers. Agent runtimes are medians rounded to the nearest minute.}
  \label{tab:time}
  \small
  \setlength{\tabcolsep}{5pt}
  \renewcommand{\arraystretch}{1.12}
  \begin{tabular}{@{}lccS[table-format=2.2]@{}}
  \toprule
  Workpaper & \makecell{Estimated manual\\workpaper time} &
  \multicolumn{1}{c}{\makecell{Median agent\\time (min)}} &
  \multicolumn{1}{c}{\makecell{Average agent\\strict pass (\%)}} \\
  \midrule
  Accounts payable & 4--6 hr & 9 & 68.18 \\
  Accounts receivable & 6--10 hr & 12 & 4.17 \\
  Cash & 3--5 hr & 8 & 56.06 \\
  Debt & 3--5 hr & 8 & 52.08 \\
  Fixed assets & 2--4 hr & 8 & 54.17 \\
  Journal-entry completeness & 2--4 hr & 7 & 94.51 \\
  Journal-entry risk testing & 6--10 hr & 12 & 16.86 \\
  Revenue & 8--12 hr & 10 & 7.58 \\
  \bottomrule
  \end{tabular}
\end{table}

\subsection{Review Tasks}
\label{app:review-results}
Financial audits follow a review process in which senior auditors review work prepared by less experienced staff. The review task setting instructs an agent to review a completed workpaper using the same supporting materials and to fix any errors that may exist. In this review setting, for each task, a completed workpaper from a randomly chosen model in the main experiments serves as the starting workpaper for every reviewing model. The results in Table~\ref{tab:review-task-results} show that all models improve average strict-pass and partial-credit scores, although some individual workpapers receive lower scores after review. Each model is run once per review task.

\begin{table}[!htbp]
\centering
\caption{Review-task results. Strict pass / partial-credit scores before and after workpaper review; changes are in percentage points. Scores before review are identical because all models receive the same starting workpapers.}
\label{tab:review-task-results}
\small
\setlength{\tabcolsep}{7pt}
\renewcommand{\arraystretch}{1.06}
\begin{tabular}{@{}lccc@{}}
\toprule
Model & \multicolumn{1}{c}{Before review (\%)} & \multicolumn{1}{c}{After review (\%)} & \multicolumn{1}{c}{Change (pp)} \\
\midrule
Claude Opus 5.5 & 28.89 / 87.34 & 62.22 / 94.57 & +33.33 / +7.23 \\
Grok 4.6 & 28.89 / 87.34 & 46.67 / 91.07 & +17.78 / +3.73 \\
GPT-5.6 Terra & 28.89 / 87.34 & 42.22 / 89.90 & +13.33 / +2.56 \\
Muse Spark 1.3 & 28.89 / 87.34 & 42.22 / 90.48 & +13.33 / +3.14 \\
DeepSeek V4.1 Flash & 28.89 / 87.34 & 37.78 / 89.59 & +8.89 / +2.25 \\
GLM-5.2 & 28.89 / 87.34 & 37.78 / 89.74 & +8.89 / +2.39 \\
GPT-5.6 Sol & 28.89 / 87.34 & 35.56 / 88.69 & +6.67 / +1.35 \\
Gemini 3.8 Flash & 28.89 / 87.34 & 33.33 / 88.67 & +4.44 / +1.33 \\
GPT-6 Astra & 28.89 / 87.34 & 33.33 / 90.37 & +4.44 / +3.03 \\
Kimi K3 & 28.89 / 87.34 & 33.33 / 89.36 & +4.44 / +2.02 \\
\bottomrule
\end{tabular}
\end{table}

\subsection{Reliability across repeated runs}
\label{app:reliability-results}
Tables~\ref{tab:pass-k} and~\ref{tab:pass-at-k} give the values plotted in Figure~\ref{fig:reliability}. Pass\textasciicircum{}k measures success on all $k$ attempts, while pass@k measures at least one success within $k$ attempts.

\begin{table}[h]
\centering
\caption{Pass\textasciicircum{}k across eight independent runs: the probability that all $k$ attempts strictly pass. Values are fractions.}
\label{tab:pass-k}
\setlength{\tabcolsep}{3.2pt}
\begin{tabular}{@{}l*{8}{S[table-format=1.2]}@{}}
\toprule
Model & \multicolumn{1}{c}{$k{=}1$} & \multicolumn{1}{c}{$k{=}2$} & \multicolumn{1}{c}{$k{=}3$} & \multicolumn{1}{c}{$k{=}4$} & \multicolumn{1}{c}{$k{=}5$} & \multicolumn{1}{c}{$k{=}6$} & \multicolumn{1}{c}{$k{=}7$} & \multicolumn{1}{c}{$k{=}8$} \\
\midrule
Claude Opus 5.5 & 0.69 & 0.60 & 0.54 & 0.50 & 0.48 & 0.46 & 0.45 & 0.44 \\
Claude Fable 5.1 & 0.66 & 0.56 & 0.49 & 0.44 & 0.40 & 0.37 & 0.34 & 0.31 \\
Grok 4.6 & 0.47 & 0.38 & 0.31 & 0.27 & 0.24 & 0.21 & 0.19 & 0.18 \\
GPT-5.6 Sol & 0.44 & 0.33 & 0.27 & 0.23 & 0.20 & 0.19 & 0.18 & 0.18 \\
GPT-6 Astra & 0.44 & 0.29 & 0.22 & 0.18 & 0.16 & 0.15 & 0.14 & 0.13 \\
GPT-5.6 Terra & 0.40 & 0.28 & 0.21 & 0.17 & 0.13 & 0.11 & 0.09 & 0.07 \\
Muse Spark 1.3 & 0.39 & 0.28 & 0.23 & 0.19 & 0.17 & 0.15 & 0.14 & 0.13 \\
DeepSeek V4.1 Flash & 0.33 & 0.21 & 0.14 & 0.10 & 0.07 & 0.05 & 0.04 & 0.02 \\
GLM-5.2 & 0.29 & 0.21 & 0.18 & 0.17 & 0.17 & 0.16 & 0.16 & 0.16 \\
Kimi K3 & 0.29 & 0.17 & 0.11 & 0.07 & 0.04 & 0.03 & 0.01 & 0.00 \\
Gemini 3.8 Flash & 0.28 & 0.22 & 0.20 & 0.19 & 0.19 & 0.18 & 0.18 & 0.18 \\
\bottomrule
\end{tabular}
\end{table}

\begin{table}[h]
\centering
\caption{Pass@k across eight independent runs: the probability of at least one strict pass within $k$ attempts.}
\label{tab:pass-at-k}
\setlength{\tabcolsep}{3.2pt}
\begin{tabular}{@{}l*{8}{S[table-format=1.2]}@{}}
\toprule
Model & \multicolumn{1}{c}{$k{=}1$} & \multicolumn{1}{c}{$k{=}2$} & \multicolumn{1}{c}{$k{=}3$} & \multicolumn{1}{c}{$k{=}4$} & \multicolumn{1}{c}{$k{=}5$} & \multicolumn{1}{c}{$k{=}6$} & \multicolumn{1}{c}{$k{=}7$} & \multicolumn{1}{c}{$k{=}8$} \\
\midrule
Claude Opus 5.5 & 0.69 & 0.79 & 0.82 & 0.84 & 0.85 & 0.85 & 0.86 & 0.87 \\
Claude Fable 5.1 & 0.66 & 0.76 & 0.79 & 0.81 & 0.82 & 0.82 & 0.82 & 0.82 \\
Grok 4.6 & 0.47 & 0.57 & 0.60 & 0.62 & 0.63 & 0.64 & 0.64 & 0.64 \\
GPT-5.6 Sol & 0.44 & 0.55 & 0.60 & 0.62 & 0.63 & 0.63 & 0.64 & 0.64 \\
GPT-6 Astra & 0.44 & 0.58 & 0.65 & 0.69 & 0.72 & 0.73 & 0.75 & 0.76 \\
GPT-5.6 Terra & 0.40 & 0.52 & 0.58 & 0.60 & 0.62 & 0.63 & 0.64 & 0.64 \\
Muse Spark 1.3 & 0.39 & 0.51 & 0.57 & 0.60 & 0.63 & 0.65 & 0.66 & 0.67 \\
DeepSeek V4.1 Flash & 0.33 & 0.45 & 0.51 & 0.55 & 0.57 & 0.58 & 0.59 & 0.60 \\
GLM-5.2 & 0.29 & 0.38 & 0.43 & 0.48 & 0.51 & 0.54 & 0.56 & 0.58 \\
Kimi K3 & 0.29 & 0.41 & 0.47 & 0.51 & 0.53 & 0.55 & 0.57 & 0.58 \\
Gemini 3.8 Flash & 0.28 & 0.33 & 0.37 & 0.39 & 0.41 & 0.43 & 0.44 & 0.44 \\
\bottomrule
\end{tabular}
\end{table}

\subsection{Effect of reasoning effort}
\label{app:reasoning-effort}
We evaluate GPT-6 Astra at five reasoning effort levels with eight independent runs per task at each reasoning effort level. Table~\ref{tab:astra-reasoning-effort} reports strict pass rates, partial-credit scores, mean runtime, and mean cost per trajectory across the five reasoning-effort levels.

\begin{table}[!htbp]
\centering
\caption{Reasoning effort ablation with GPT-6 Astra. Runtime and model cost are means per trajectory, with eight independent runs per task.}
\label{tab:astra-reasoning-effort}
\setlength{\tabcolsep}{5pt}
\renewcommand{\arraystretch}{1.08}
\begin{tabular}{@{}l*{4}{S[table-format=2.2]}@{}}
\toprule
Reasoning effort & \multicolumn{1}{c}{Strict pass (\%)} & \multicolumn{1}{c}{Partial credit (\%)} & \multicolumn{1}{c}{\makecell{Mean runtime\\(min)}} & \multicolumn{1}{c}{\makecell{Mean cost\\(\$)}} \\
\midrule
Low & 23.61 & 85.71 & 3.08 & 1.32 \\
Medium & 43.61 & 90.95 & 8.00 & 2.96 \\
High & 55.00 & 92.71 & 9.59 & 3.97 \\
Xhigh & 59.44 & 94.16 & 10.95 & 4.49 \\
Max & 71.11 & 95.83 & 24.94 & 8.27 \\
\bottomrule
\end{tabular}
\end{table}

\subsection{Experimental and grading details}
\paragraph{Model configurations.}
Table~\ref{tab:model-configurations} specifies the settings used in the main experiments of this work. Reasoning effort is configured for each model and set to the default or recommended level reported in its corresponding model card. For Fable 5.1, we set the reasoning effort to medium to be more in line with other models we evaluate. We used the default API endpoints for all OpenAI and Anthropic models and OpenRouter for the remaining models. The default caching policy is used for all models. Other parameters not explicitly outlined here retain their default values.

\begin{table}[h!]
\centering
\caption{Models and inference settings for the main and review experiments. Note that Claude Fable 5.1 is evaluated only in the main experiments.}
\label{tab:model-configurations}
\setlength{\tabcolsep}{3pt}
\renewcommand{\arraystretch}{1.15}
\begin{tabular}{@{}llcS[table-format=1.1]@{}}
\toprule
Model & API & \makecell{Reasoning\\effort} & \multicolumn{1}{c}{Temperature} \\
\midrule
Claude Opus 5.5 & \texttt{claude-opus-5-5} & medium & {--} \\
Claude Fable 5.1 & \texttt{claude-fable-5-1} & medium & {--} \\
Grok 4.6 & \texttt{x-ai/grok-4.6} & medium & {--} \\
GPT-5.6 Sol & \texttt{gpt-5.6-sol} & medium & {--} \\
GPT-6 Astra & \texttt{gpt-6-astra} & medium & {--} \\
GPT-5.6 Terra & \texttt{gpt-5.6-terra} & medium & {--} \\
Muse Spark 1.3 & \texttt{meta/muse-spark-1.3} & medium & {--} \\
DeepSeek V4.1 Flash & \texttt{deepseek/deepseek-v4.1-flash} & medium & {--} \\
GLM-5.2 & \texttt{z-ai/glm-5.2} & high & {--} \\
Kimi K3 & \texttt{moonshotai/kimi-k3} & medium & 1.0 \\
Gemini 3.8 Flash & \texttt{google/gemini-3.8-flash} & medium & {--} \\
\bottomrule
\end{tabular}
\end{table}

\paragraph{Experimental setting.}
In the main experimental setting, each model-task pair is run independently eight times. Every run receives the same system prompt, task instructions, client file room, and blank workpaper template in an isolated Docker workspace. The agent can interact with the workspace only through a \textit{bash} tool. A run ends when the agent returns the required completion signal or reaches 200 model calls. Individual model requests have a 600-second timeout and up to two retries. The agent harness does not perform context compaction or summarization. Less than one percent of runs failed due to external errors such as API errors, and we reran those trajectories.

\paragraph{Performance by workpaper area.}
Table~\ref{tab:workpaper-area-results} reports partial-credit scores by workpaper area. For each model, reported partial-credit scores are within each area. This breakdown shows how aggregate model performance varies across the different audit areas we evaluate.

\begin{table}[h!]
\centering
\caption{Partial-credit scores by workpaper area (\%).}
\label{tab:workpaper-area-results}
\scriptsize
\setlength{\tabcolsep}{2.1pt}
\renewcommand{\arraystretch}{1.25}
\newcommand{\angledmodel}[1]{\makebox[0pt][l]{\rotatebox{45}{\strut #1}}}
\begin{tabular*}{\linewidth}{@{\extracolsep{\fill}}lS[table-format=3.2]S[table-format=2.2]S[table-format=2.2]S[table-format=3.2]S[table-format=3.2]S[table-format=2.2]S[table-format=2.2]S[table-format=2.2]S[table-format=2.2]S[table-format=2.2]S[table-format=3.2]@{\hspace{38pt}}}
Workpaper area & \multicolumn{1}{c}{\angledmodel{Claude Opus 5.5}} & \multicolumn{1}{c}{\angledmodel{Claude Fable 5.1}} & \multicolumn{1}{c}{\angledmodel{Grok 4.6}} & \multicolumn{1}{c}{\angledmodel{GPT-5.6 Sol}} & \multicolumn{1}{c}{\angledmodel{GPT-6 Astra}} & \multicolumn{1}{c}{\angledmodel{GPT-5.6 Terra}} & \multicolumn{1}{c}{\angledmodel{Muse Spark 1.3}} & \multicolumn{1}{c}{\angledmodel{DeepSeek V4.1 Flash}} & \multicolumn{1}{c}{\angledmodel{GLM-5.2}} & \multicolumn{1}{c}{\angledmodel{Kimi K3}} & \multicolumn{1}{c@{\hspace{38pt}}}{\angledmodel{Gemini 3.8 Flash}} \\
\midrule
Accounts payable & 100.00 & 98.96 & 99.48 & 98.03 & 94.44 & 96.06 & 98.44 & 90.57 & 91.44 & 94.04 & 92.77 \\
Accounts receivable & 89.43 & 87.01 & 81.30 & 83.78 & 80.63 & 79.47 & 84.58 & 80.23 & 80.53 & 84.08 & 81.20 \\
Cash & 99.00 & 99.43 & 97.66 & 96.16 & 93.67 & 96.80 & 94.09 & 95.21 & 91.42 & 88.06 & 89.45 \\
Debt & 99.85 & 99.08 & 96.38 & 94.35 & 92.15 & 95.50 & 94.50 & 94.60 & 91.15 & 89.68 & 95.11 \\
Fixed assets & 95.82 & 94.82 & 95.49 & 89.42 & 88.21 & 93.39 & 95.87 & 93.70 & 94.94 & 92.58 & 87.64 \\
Journal-entry completeness & 100.00 & 99.58 & 99.65 & 100.00 & 100.00 & 96.04 & 99.24 & 96.46 & 99.58 & 93.19 & 100.00 \\
Journal-entry risk testing & 97.08 & 96.59 & 84.53 & 82.02 & 91.61 & 78.69 & 81.67 & 82.01 & 82.91 & 81.91 & 83.50 \\
Revenue & 87.60 & 85.07 & 78.67 & 75.81 & 88.63 & 77.03 & 72.98 & 74.82 & 70.47 & 68.52 & 75.83 \\
\midrule
\textbf{Overall} & \bfseries 95.84 & \bfseries 94.81 & \bfseries 91.12 & \bfseries 89.41 & \bfseries 90.95 & \bfseries 88.66 & \bfseries 89.62 & \bfseries 88.31 & \bfseries 87.56 & \bfseries 86.01 & \bfseries 87.88 \\
\bottomrule
\end{tabular*}
\end{table}

\paragraph{Rubrics and grading.}
Prior approaches to grading work completed in spreadsheets have included methods such as extracting and checking exact cell coordinates~\citep{ma2024spreadsheetbenchchallengingrealworld} and evaluating completed work with LLMs~\citep{yen2026mbabenchevaluatingllmagents}. In this work, rubrics are written by auditors who check the presence, absence, and/or correctness of specific cells or rows deemed vital in each workpaper. As the purpose of these workpapers is to both collect \textit{and} document evidence of the audit procedure being carried out, our rubric criteria are designed to accommodate substantively equivalent answers rather than being overly restrictive as different auditors may apply their own preferences for grading dimensions such as how much documentation is sufficient. Grading criteria range from checking that particular transactions are documented and recorded correctly to verifying that the correct conclusion(s) are reached.

The grading procedure implemented in this work leverages the fact that each worksheet in each of our workpaper templates consists of one or more vertically stacked tables with fixed columns. Agents may add or remove rows from each table during each task. Yet, they are instructed not to make changes to the columns of the templates themselves. Our provided rubric criteria are then applied to each table via a keyed approach that first locates the column of the table to be evaluated and then extracts values from target row(s). This allows grading to accommodate changes in row positions without relying on fixed cell coordinates. When multiple valid ways of completing a table make the relevant rows or cells difficult to identify programmatically, the grading procedure passes the relevant table content to the grader for evaluation. \textit{GPT-5.6 Luna} with medium reasoning effort is used as the model for LLM-based evaluation.

\paragraph{Rubric Construction.}
To facilitate efficient rubric construction and revisions and to provide visibility into how workpapers are graded, the evaluation code highlights cells that are evaluated and subsequently graded as correct or incorrect with different colors. Furthermore, each graded cell is highlighted and annotated with its corresponding grading outcome and reasoning. In practice, these annotations allow auditors to more efficiently review and revise rubrics and researchers to understand and analyze what types of errors models make. LLMs were used to translate feedback provided by auditors on the rubrics into evaluation code.

\begin{figure}[htbp]
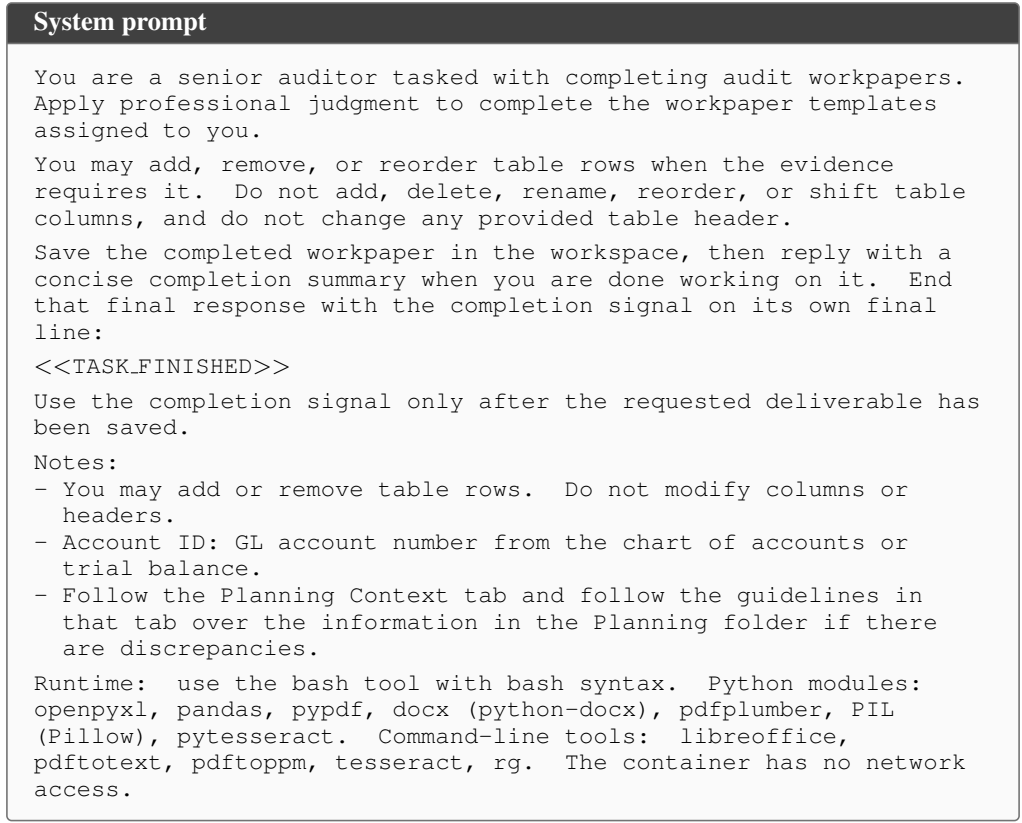

\centering
  \begin{tcolorbox}[
    width=0.96\linewidth,
    colback=black!2,
    colframe=black!55,
    colbacktitle=black!75,
    coltitle=white,
    title={System prompt},
    fonttitle=\bfseries,
    fontupper=\ttfamily\footnotesize\raggedright,
    boxrule=0.6pt,
    arc=1.5pt,
    left=7pt,
    right=7pt,
    top=6pt,
    bottom=6pt
  ]
You are a senior auditor tasked with completing audit workpapers. Apply professional judgment to complete the workpaper templates assigned to you.
\par\smallskip
You may add, remove, or reorder table rows when the evidence requires it. Do not add, delete, rename, reorder, or shift table columns, and do not change any provided table header.
\par\smallskip
Save the completed workpaper in the workspace, then reply with a concise completion summary when you are done working on it. End that final response with the completion signal on its own final line:
\par\smallskip
\textless{}\textless{}TASK\_FINISHED\textgreater{}\textgreater{}
\par\smallskip
Use the completion signal only after the requested deliverable has been saved.
\par\smallskip
Notes:
\par
{\hangindent=1.2em\hangafter=1 - You may add or remove table rows. Do not modify columns or headers.\par}
{\hangindent=1.2em\hangafter=1 - Account ID: GL account number from the chart of accounts or trial balance.\par}
{\hangindent=1.2em\hangafter=1 - Follow the Planning Context tab and follow the guidelines in that tab over the information in the Planning folder if there are discrepancies.\par}
\par\smallskip
Runtime: use the bash tool with bash syntax. Python modules: openpyxl, pandas, pypdf, docx (python-docx), pdfplumber, PIL (Pillow), pytesseract. Command-line tools: libreoffice, pdftotext, pdftoppm, tesseract, rg. The container has no network access.
  \end{tcolorbox}
\caption{System prompt provided to the agent in the preparation and review settings.}
\label{fig:system-prompt}
\end{figure}

\paragraph{System prompt.}
Figure~\ref{fig:system-prompt} shows the system prompt provided to our agent in both the preparation and review settings. The notes also clarify what Account ID refers to, as some models commonly confused that terminology. The computer-use experiment in particular uses the system prompt supplied by the ChatGPT application. Figure~\ref{fig:computer-use-instructions} shows the instructions supplied as a user prompt for this experiment.

\paragraph{Code Release.}
We release a \textit{development version} of the task generation framework that preserves the overall structure of the code while omitting much of the domain-specific logic. This version is intended to make the framework easier for the broader research community to understand. The complete benchmark tasks and evaluation code are released in their entirety.

\begin{figure}[htbp]
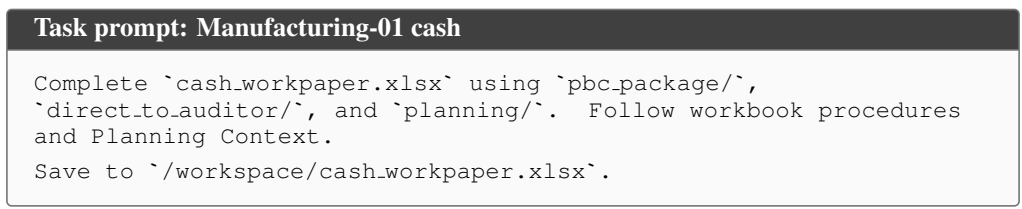

\centering
  \begin{tcolorbox}[
    width=0.96\linewidth,
    colback=black!2,
    colframe=black!55,
    colbacktitle=black!75,
    coltitle=white,
    title={Task prompt: Manufacturing-01 cash},
    fonttitle=\bfseries,
    fontupper=\ttfamily\footnotesize\raggedright,
    boxrule=0.6pt,
    arc=1.5pt,
    left=7pt,
    right=7pt,
    top=6pt,
    bottom=6pt
  ]
Complete \textasciigrave{}cash\_workpaper.xlsx\textasciigrave{} using \textasciigrave{}pbc\_package/\textasciigrave{}, \textasciigrave{}direct\_to\_auditor/\textasciigrave{}, and \textasciigrave{}planning/\textasciigrave{}. Follow workbook procedures and Planning Context.
\par\smallskip
Save to \textasciigrave{}/workspace/cash\_workpaper.xlsx\textasciigrave{}.
  \end{tcolorbox}
\caption{Example task prompt for a workpaper completion task.}
\label{fig:task-instructions}
\end{figure}

\paragraph{Instructions.}
Figure~\ref{fig:task-instructions} shows an example of the user instructions provided at the start of a workpaper-preparation task.

\begin{figure}[htbp]
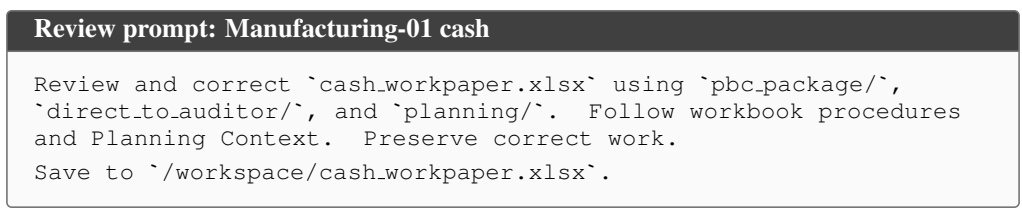

\centering
  \begin{tcolorbox}[
    width=0.96\linewidth,
    colback=black!2,
    colframe=black!55,
    colbacktitle=black!75,
    coltitle=white,
    title={Review prompt: Manufacturing-01 cash},
    fonttitle=\bfseries,
    fontupper=\ttfamily\footnotesize\raggedright,
    boxrule=0.6pt,
    arc=1.5pt,
    left=7pt,
    right=7pt,
    top=6pt,
    bottom=6pt
  ]
Review and correct \textasciigrave{}cash\_workpaper.xlsx\textasciigrave{} using \textasciigrave{}pbc\_package/\textasciigrave{}, \textasciigrave{}direct\_to\_auditor/\textasciigrave{}, and \textasciigrave{}planning/\textasciigrave{}. Follow workbook procedures and Planning Context. Preserve correct work.
\par\smallskip
Save to \textasciigrave{}/workspace/cash\_workpaper.xlsx\textasciigrave{}.
  \end{tcolorbox}
\caption{Example task prompt for a workpaper review task.}
\label{fig:review-instructions}
\end{figure}

\paragraph{Review instructions.}
Figure~\ref{fig:review-instructions} shows the corresponding user instructions in the review setting. The agent receives a previously completed workpaper and the same supporting materials from the engagement binder, with instructions to correct errors while preserving correct work.

\begin{figure}[htbp]
\centering
\includegraphics[width=\linewidth]{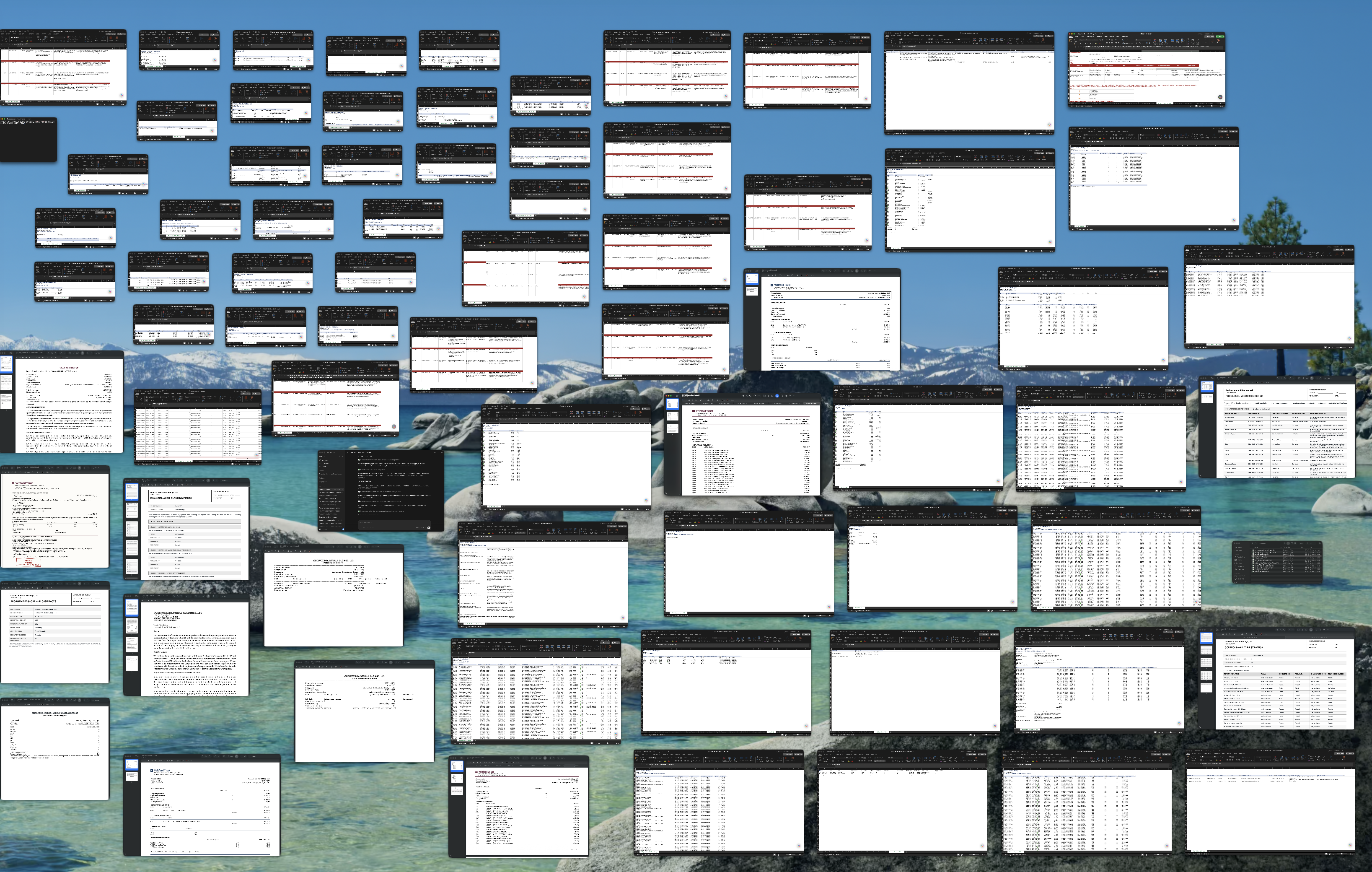}
\caption{Final desktop overview during the computer-use experiment with ChatGPT Computer-Use.}
\label{fig:computer-use-open-windows}
\end{figure}

\paragraph{Computer-use instructions.}
Figure~\ref{fig:computer-use-instructions} shows the task instructions supplied as a user prompt to ChatGPT for one of our synthetic binders. The desktop application took 283 minutes to complete the eight workpapers, compared with a mean total execution time of 65 minutes for the bash-based agent on the same tasks. During the computer-use run, ChatGPT left document windows open after use, making its activity difficult to track (Figure~\ref{fig:computer-use-open-windows}).

\begin{figure}[htbp]
\centering
  \begin{tcolorbox}[
    width=0.96\linewidth,
    colback=black!2,
    colframe=black!55,
    colbacktitle=black!75,
    coltitle=white,
    title={Computer-use prompt},
    fonttitle=\bfseries,
    fontupper=\ttfamily\footnotesize\raggedright,
    boxrule=0.6pt,
    arc=1.5pt,
    left=7pt,
    right=7pt,
    top=6pt,
    bottom=6pt
  ]
You are a senior auditor tasked with completing audit workpapers.
\par
Apply professional judgment to complete the workpaper templates assigned to you.
\par\smallskip
Binder folder:
\par
\textless{}binder-folder\textgreater{}
\par\smallskip
Complete every workpaper in the binder folder’s templates/ directory using the supporting materials in:
\par
- data\_room/pbc\_package/
\par
- data\_room/direct\_to\_auditor/
\par
- data\_room/planning/
\par\smallskip
Use computer-use tools to navigate the folders, read the supporting documents, and edit the workbooks in desktop applications.
\par\smallskip
Access only the designated binder folder and its subfolders. Do not read, search, or modify files outside that folder, access other binders, or use websites or external data sources.
\par\smallskip
Follow each workbook’s procedures and Planning Context tab. If the Planning Context tab conflicts with documents in data\_room/planning/, follow the Planning Context tab.
\par\smallskip
You may add, remove, or reorder table rows when the evidence requires it. Do not add, delete, rename, reorder, or shift table columns, and do not change any provided table header.
\par\smallskip
“Account ID” means the GL account number from the chart of accounts or trial balance.
\par\smallskip
Save each completed workbook in templates/ under its original filename. Continue until you have completed and saved all assigned workpapers.
\par\smallskip
Then provide a concise completion summary.
  \end{tcolorbox}
\caption{User prompt for completing all workpapers in a binder with computer-use tools. The local binder-folder path is replaced with a placeholder.}
\label{fig:computer-use-instructions}
\end{figure}

\FloatBarrier

\end{document}

%% file: math_commands.tex
\usepackage{amsmath,amsfonts,bm}

\def\eqref#1{equation~\ref{#1}}

\def\1{\bm{1}}

\DeclareMathAlphabet{\mathsfit}{\encodingdefault}{\sfdefault}{m}{sl}
\SetMathAlphabet{\mathsfit}{bold}{\encodingdefault}{\sfdefault}{bx}{n}

